\documentclass[10pt,twocolumn,letterpaper]{article}
\usepackage{cvpr}
\usepackage{times}
\usepackage{epsfig}
\usepackage{graphicx}
\usepackage{amsmath}
\usepackage{amssymb}
\usepackage{tabularx}
\usepackage{subcaption}

\usepackage[pagebackref=true,breaklinks=true,letterpaper=true,colorlinks,bookmarks=false]{hyperref}

\newcommand{\bx}{\mathbf{x}}

\newcommand{\bw}{\mathbf{w}}

\newcommand{\bv}{\mathbf{v}}
\newcommand{\bmf}{\mathbf{f}}

\usepackage{xspace}
\makeatletter
\DeclareRobustCommand\onedot{\futurelet\@let@token\@onedot}
\def\@onedot{\ifx\@let@token.\else.\null\fi\xspace}

\def\eg{\emph{e.g}\onedot} 
\def\ie{\emph{i.e}\onedot}

\def\etal{\emph{et al}\onedot}
\makeatother

\usepackage{xcolor}

\newcommand{\red}[1]{\textcolor{red}{#1}}
\newcommand{\green}[1]{\textcolor{green}{#1}}

\cvprfinalcopy 

\def\cvprPaperID{3213} 

\ifcvprfinal\fi
\begin{document}

\title{ Unsupervised Embedding Learning via Invariant and Spreading \\ Instance Feature \vspace{-0.2cm}}

\author{Mang Ye$^{\dag}$ \quad Xu Zhang$^{\ddag}$ \quad Pong C. Yuen$^{\dag}$ \quad  Shih-Fu Chang$^{\ddag}$\vspace{0.2cm}\\
$^{\dag}$ Hong Kong Baptist University, Hong Kong \quad \quad $^{\ddag}$ Columbia University, New York \\
{\tt\small \{mangye,pcyuen\}@comp.hkbu.edu.hk, \{xu.zhang,sc250\}@columbia.edu}
}

\maketitle
\thispagestyle{empty}

\begin{abstract}
This paper studies the unsupervised embedding learning problem, which requires an effective similarity measurement between samples in low-dimensional embedding space.
Motivated by the positive concentrated and negative separated properties observed from category-wise supervised learning,
we propose to utilize the instance-wise supervision to approximate these properties, which aims at learning data augmentation invariant and instance spread-out features.
To achieve this goal, we propose a novel instance based softmax embedding method, which directly optimizes the `real' instance features on top of the softmax function.
It achieves significantly faster learning speed and higher accuracy than all existing methods.
The proposed method performs well for both seen and unseen testing categories with cosine similarity. It also achieves competitive performance even without pre-trained network over samples from fine-grained categories.
\end{abstract}

\section{Introduction}
\label{sec:intro}
Deep embedding learning is a fundamental task in computer vision~\cite{iccv15img}, which aims at learning a feature embedding that has the following properties:
1) \textit{positive concentrated}, the embedding features of samples belonging to the same category are close to each other~\cite{cvpr16lifted};
2) \textit{negative separated}, the embedding features of samples belonging to different categories are separated as much as possible~\cite{iccv17spread}.
Supervised embedding learning methods have been studied to achieve such objectives and demonstrate impressive capabilities in various vision tasks ~\cite{aaai15surpass,iccv17nca,arxiv17align}.
However, annotated data needed for supervised methods might be difficult to obtain. Collecting enough annotated data for different tasks requires costly human efforts and special domain expertise.
To address this issue, this paper tackles the unsupervised embedding learning problem (a.k.a. unsupervised metric learning in~\cite{cvpr18mom}), which aims at learning discriminative embedding features without human annotated labels.

Unsupervised embedding learning usually requires that the similarity between learned embedding features is consistent with the visual similarity or category relations of input images.
In comparison, general unsupervised feature learning usually aims at learning a good ``intermediate" feature representation from unlabelled data~\cite{iccv15context,nips16cluster,eccv16jigsaw,cvpr16context}. The learned feature is then generalized to different tasks by using a small set of labelled training data from the target task to fine-tune models (\eg, linear classifier, object detector, etc.) for the target task \cite{eccv18cluster}.
However, the learned feature representation may not preserve visual similarity and its performance drops dramatically for similarity based tasks, \eg nearest neighbor search~\cite{cvpr18nce,eccv18race,iccv17dgm}.
\begin{figure}[t]
  \centering
  \includegraphics[width = 8cm]{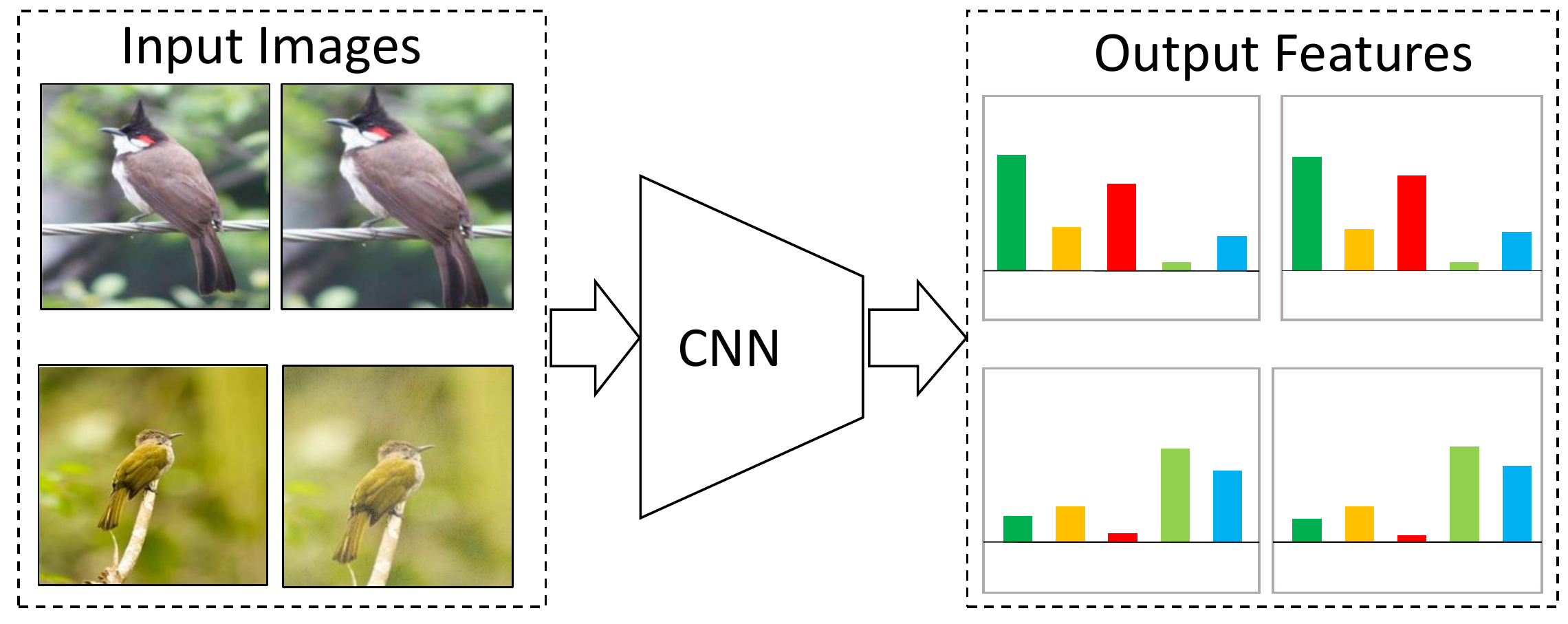}\\
  \caption{\small{Illustration of our basic idea. The features of the same instance under different data augmentations should be invariant, while features of different image instances should be separated.}}\label{fig:idea}
\end{figure}
%


The main challenge of unsupervised embedding learning is to discover visual similarity or weak category information from unlabelled samples.
Iscen \etal~\cite{cvpr18mom} proposed to mine hard positive and negative samples on manifolds.
However, its performance heavily relies on the quality of the initialized feature representation for label mining, which limits the applicability for general tasks.
In this paper, we propose to utilize the instance-wise supervision to approximate the \textit{positive concentrated} and \textit{negative separated} properties mentioned earlier.
The learning process only relies on instance-wise relationship and does not rely on relations between pre-defined categories, so it can be well generalized to samples of arbitrary categories that have not been seen before (\textit{unseen testing categories})~\cite{cvpr06invariant}.
%
%

\textit{For positive concentration:} it is usually infeasible to mine reliable positive information with randomly initialized network. Therefore,
we apply a random data augmentation (\eg, transformation, scaling) to each image instance and use the augmented image as a positive sample.
In other words, features of each image instance under different data augmentations should be invariant.
\textit{For negative separation}: since unlabelled data are usually highly imbalanced \cite{pami16noise,tip19dgm}, the number of negative samples for each image instance is much larger than that of positive samples.
Therefore, a small batch of randomly selected instances can be approximately treated as negative samples
for each instance. With such assumption, we try to separate each instance from all the other sampled instances within the batch, resulting in a spread-out property~\cite{iccv17spread}. It is clear that such assumption may not always hold, and each batch may contain a few
false negatives. However, through our extensive experiments, we observe that the spread-out property effectively improves the discriminability.
%
%
%
In summary, our main idea is to learn a discriminative instance feature, which preserves \textit{data augmentation invariant} and \textit{spread-out} properties for unsupervised embedding learning, as shown in Fig. \ref{fig:idea}.


To achieve these goals, we introduce a novel instance feature-based softmax embedding method.
Existing softmax embedding is usually built on classifier weights \cite{pami16exampler} or memorized features \cite{cvpr18nce}, which has limited efficiency and discriminability.
We propose to explicitly optimize the feature embedding by directly using the inner products of instance features on top of softmax function, leading to significant performance and efficiency gains. The softmax function mines hard negative samples and takes full advantage of relationships among all sampled instances to improve the performance. 
%
The number of instance is significantly larger than the number of categories, so we introduce a Siamese network training strategy. We transform the
multi-class classification problem to a binary classification problem and use maximum likelihood estimation for optimization.

The main contributions can be summarized as follows:
\begin{itemize}
\setlength\itemsep{0.1em}
\item We propose a novel instance feature-based softmax embedding method to learn data augmentation invariant and instance spread-out features. It achieves significantly faster learning speed and higher accuracy than all the competing methods.
%
\item We show that both the data augmentation invariant and instance spread-out properties are important for instance-wise unsupervised embedding learning. They help capture apparent visual similarity between samples and generalizes well on unseen testing categories.
\item The proposed method achieves the state-of-the-art performances over other unsupervised learning methods on comprehensive image classification and embedding learning experiments.
\end{itemize}

\section{Related Work}
\textbf{General Unsupervised Feature Learning.}
Unsupervised feature learning has been widely studied in literature. Existing works can be roughly categorized into three categories~\cite{eccv18cluster}:
1) \textit{generative models}, this approach aims at learning a parameterized mapping between images and predefined noise signals, which constrains the distribution between raw data and noises~\cite{cvpr18nce}. Bolztmann Machines (RBMs)~\cite{icml09rbm,cvpr12rbm}, Auto-encoders~\cite{cvpr07autoencoder,icml08autoencoder} and generative adversarial network (GAN)~\cite{arxiv16gan,arxiv16ali,nips14gan} are widely studied.
2) \textit{Estimating Between-image Labels,} it usually estimates between-image labels using the clustering technique \cite{eccv18cluster,nips14cluster,nips16cluster} or kNN-based methods \cite{icml13knn}, which provide label information. Then label information and feature learning process are iteratively updated.
3) \textit{Self-supervised Learning,} this approach designs pretext tasks/signals to generate ``pseudo-labels" and then formulate it as a prediction task to learn the feature representations. The pretext task could be the context information of local patches \cite{iccv15context}, the position of randomly rearranged patches \cite{eccv16jigsaw}, the missing pixels of an image \cite{cvpr16context} or the color information from gray-scale images \cite{eccv16color}. Some attempts also use video information to provide weak supervision to learn feature representations \cite{iccv15move,iccv15video}.

As we discussed in Section~\ref{sec:intro}, general unsupervised feature learning usually aims at learning a good ``intermediate" feature representation that can be well generalized to other tasks. The intermediate feature representation may not preserve visual similar property. In comparison, unsupervised embedding learning requires additional visual similarity property of the learned features.


\textbf{Deep Embedding Learning.} Deep embedding learning usually learns an embedding function by minimizing the intra-class variation and maximizing the inter-class variation \cite{cvpr16lifted,nips16npair,eccv16deep,cvpr16dgd}. Most of them are  designed on top of pairwise \cite{cvpr06invariant,iccv17nca} or triplet relationships \cite{iccv17smart,iccv17embed}. In particular, several sampling strategies are widely investigated to improve the performance, such as hard mining \cite{arxiv17triplet}, semi-hard mining \cite{cvpr15face}, smart mining \cite{iccv17smart} and so on. In comparison, softmax embedding achieves competitive performance without sampling requirement \cite{arxiv17softmax}. Supervised learning has achieved superior performance on various tasks, but they still rely on enough annotated data.

\textbf{Unsupervised Embedding Learning.}  According to the evaluation protocol, it can be categorized into two cases, 1) the testing categories are the same with the training categories (\textit{seen testing categories}), and 2) the testing categories are not overlapped with the training categories (\textit{unseen testing categories}). The latter setting is more challenging. Without category-wise labels, Iscen \textit{et al.} \cite{cvpr18mom} proposed to mine hard positive and negative samples on manifolds, and then train the feature embedding with triplet loss. However, it heavily relies on the initialized representation for label mining.


\section{Proposed Method}
Our goal is to learn a feature embedding network $f_\theta(\cdot)$ from a set of unlabelled images $X=\{\mathbf{x}_1, \mathbf{x}_2, \cdots, \mathbf{x}_n\}$. $f_\theta(\cdot)$ maps the input image $\mathbf{x}_i$ into a low-dimensional embedding feature $f_\theta(\mathbf{x}_i) \in \mathbb{R}^{d}$, where $d$ is the feature dimension. For simplicity, the feature representation $f_\theta(\mathbf{x}_i)$ of an image instance is represented by $\mathbf{f}_i$, and we assume that all the features are $\ell_2$ normalized, \ie $\|\bmf_i \|_2 =1$.
A good feature embedding should satisfy:
1) the embedding features of visual similar images are close to each other;
2) the embedding features of dissimilar image instances are separated.

Without category-wise labels, we utilize the instance-wise supervision to approximate the \textit{positive concentrated} and \textit{negative seperated} properties.
In particular, the embedding features of the same instance under different data augmentations should be invariant, while the features of different instances should be spread-out.
In the rest of this section, we first review two existing instance-wise feature learning methods, and then propose a much more efficient and discriminative instance feature-based softmax embedding.
Finally, we will give a detailed rationale analysis and introduce our training strategy with Siamese network.
\subsection{Instance-wise Softmax Embedding}
\noindent\textbf{Softmax Embedding with Classifier Weights.}
%
Exemplar CNN \cite{pami16exampler} treats each image as a distinct class. Following the conventional classifier training, it defines a matrix $\mathbf{W} = [\mathbf{w}_1, \mathbf{w}_2, \cdots, \mathbf{w}_n ]^T \in \mathbb{R}^{n\times d}$, where the $j$-th column $\mathbf{w}_j$ is called the corresponding classifier weight for the $j$-th instance.
Exemplar CNN ensures that image instance under different image transformations can be correctly classified into its original instance with the learned weight.
%
%
Based on Softmax function, the probability of sample $\mathbf{x}_j$ being recognized as the $i$-th instance can be represented as
\begin{equation}\label{eq:wsoftmax}
P(i|\mathbf{x}_j) = \frac{\exp (\mathbf{w}_i^T\mathbf{f}_j)}{\sum\nolimits_{k = 1}^n \exp (\mathbf{w}_k^T\mathbf{f}_j)}.
\end{equation}
At each step, the network pulls sample feature $\bmf_i$ towards its corresponding weight $\mathbf{w}_i$, and pushes it away from the classifier weights $\mathbf{w}_k$ of other instances. However, classifier weights prevent explicitly comparison over features, which results in limited efficiency and discriminability.

\noindent\textbf{Softmax Embedding with Memory Bank.}
To improve the inferior efficiency, Wu \etal~\cite{cvpr18nce} propose to set up a memory bank to store the instance features $\bmf_i$ calculated in the previous step.
The feature stored in the memory bank is denoted as $\bv_i$, which serves as the classifier weight for the corresponding instance in the following step.
Therefore, the probability of sample $\bx_j$ being recognized as the $i$-th instance can be written as
\begin{equation}
\label{eq:msoftmax}
P(i|\bx_j) = \frac{\exp(\bv_i^T\bmf_j / \tau)}{\sum\nolimits_{k = 1}^n \exp (\bv_k^T\bmf_j / \tau)},
\end{equation}
where $\tau$ is the temperature parameter controlling the concentration level of the sample distribution \cite{arxiv15temp}.
$\mathbf{v}_i^T{\mathbf{f}_j}$ measures the cosine similarity between the feature $\bmf_j$ and the $i$-th memorized feature $\mathbf{v}_i$. For instance $\mathbf{x}_i$ at each step, the network pulls its feature $\bmf_i$ towards its corresponding memorized vector $\mathbf{v}_i$, and pushes it away from the memorized vectors of other instances. Due to efficiency issue, the memorized feature $\mathbf{v}_i$ corresponding to instance $\mathbf{x}_i$ is only updated in the iteration which takes $\mathbf{x}_i$ as input. In other words, the memorized feature $\mathbf{v}_i$ is only updated once per epoch.
However, the network itself is updated in each iteration. Comparing the real-time instance feature $\bmf_i$ with the outdated memorized feature $\mathbf{v}_i$ would cumber the training process. Thus, the memory bank scheme is still inefficient.

A straightforward idea to improve the efficiency is directly optimizing over feature itself, \ie replacing the weight $\{\bw_i\}$ or memory $\{\bv_i\}$ with $\bmf_i$. However, it is implausible due to two reasons:
1) Considering the probability $P(i|\bx_i)$ of recognizing $\bx_i$ to itself, since $\bmf_i^T\bmf_i$=1, \ie the feature and ``pseudo classifier weight'' (the feature itself) are always perfectly aligned, optimizing the network will not provide any positive concentrated property;
2) It's impractical to calculate the feature of all the samples ($\bmf_k, k = 1,\ldots,n$) on-the-fly in order to calculate the denominator in Eq.~(\ref{eq:msoftmax}), especially for large-scale instance number dataset.

\subsection{Softmax Embedding on `Real' Instance Feature}
To address above issues, we propose a softmax embedding variant for unsupervised embedding learning, which directly optimizes the real instance feature rather than classifier weights \cite{pami16exampler} or memory bank \cite{cvpr18nce}. To achieve the goal that features of the same instance under different data augmentations are invariant, while the features of different instances are spread-out,
we propose to consider 1) both the original image and its augmented image, 2) a small batch of randomly selected samples instead of the full dataset.

For each iteration, we randomly sample $m$ instances from the dataset. To simplify the notation, without loss of generality, the selected samples are denoted by $\{\bx_1, \bx_2, \cdots, \bx_m\}$. For each instance, a random data augmentation operation $T(\cdot)$ is applied to slightly modify the original image. The augmented sample $T(\bx_i)$ is denoted by $\hat{\bx}_i$, and its embedding feature $f_\theta(\hat{\bx}_i)$ is denoted by $\hat{\bmf}_i$.
Instead of considering the instance feature learning as a multi-class classification problem, we solve it as binary classification problem via maximum likelihood estimation (MLE). In particular, for instance $\bx_i$, the augmented sample $\hat{\bx}_i$ should be classified into instance $i$, and other instances $\bx_j, \ j \neq i$ shouldn't be classified into instance $i$.
\begin{figure*}[t]
  \centering
  \includegraphics[ width = 17cm]{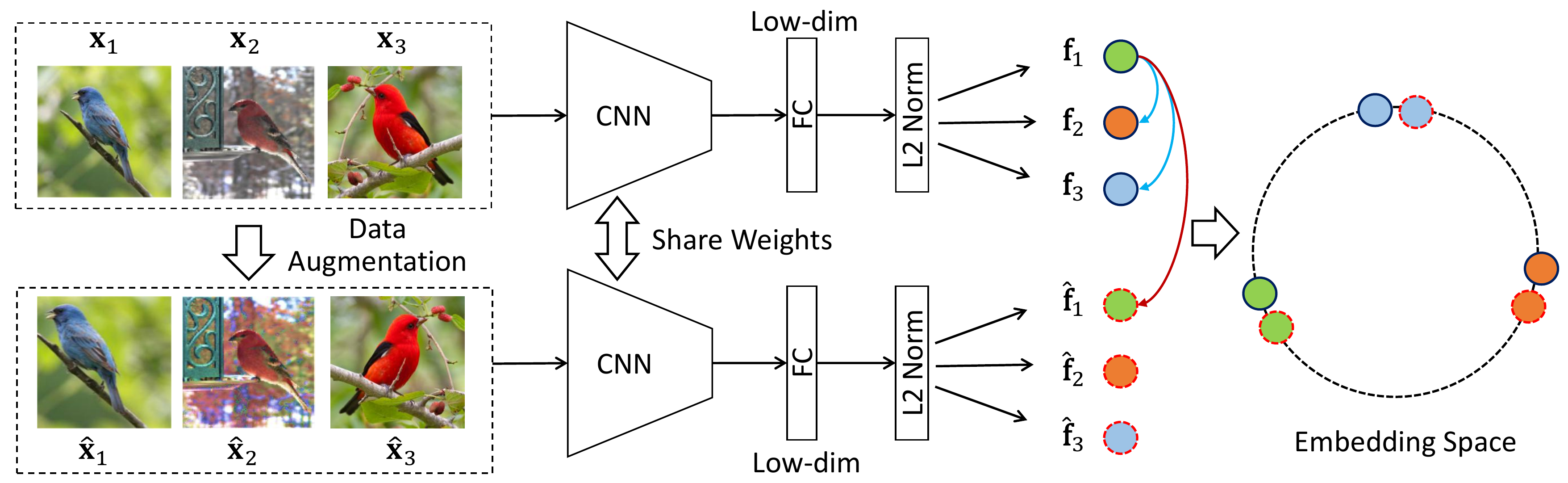}\\
  \caption{\small{The framework of the proposed unsupervised learning method with Siamese network. The input images are projected into low-dimensional normalized embedding features with the CNN backbone. Image features of the same image instance with different data augmentations are invariant, while embedding features of different image instances are spread-out.} }\label{fig:framework}
  \vspace{-0.3cm}
\end{figure*}
The probability of $\hat{\bx}_i$ being recognized as instance $i$ is defined by
\begin{equation}\label{eq:pos}
  P(i|\hat{\bx}_i) = \frac{\exp(\bmf_i^T\hat{\bmf}_i/\tau)}{\sum\nolimits_{k=1}^m \exp (\bmf_k^T\hat{\bmf}_i/\tau)}.
\end{equation}
On the other hand, the probability of $\bx_j$ being recognized as instance $i$ is defined by
\begin{equation}\label{eq:neg}
  P(i|\bx_j) = \frac{\exp(\bmf_i^T\bmf_j/\tau)}{\sum\nolimits_{k=1}^m \exp (\bmf_k^T\bmf_j/\tau)}, \ j \neq i
\end{equation}
Correspondingly, the probability of $\bx_j$ \emph{not} being recognized as instance $i$ is $1-P(i|\bx_j)$.

Assuming different instances being recognized as instance $i$ are independent, the joint probability of $\hat{\bx}_i$ being recognized as instance $i$ and $\bx_j, \ j \neq i$ not being classified into instance $i$ is
\begin{equation}\label{eq:jointP}
  P_{i} = P(i|\hat{\bx}_i)\prod_{j \neq i}(1-P(i|\bx_j))
\end{equation}
The negative log likelihood is given by
\begin{equation}\label{eq:log}
  J_{i} = -\log P(i|\hat{\bx}_i) - \sum_{j \neq i}\log(1-P(i|\bx_j))
\end{equation}
We solve this problem by minimizing the sum of the negative log likelihood over all the instances within the batch, which is denoted by
\begin{equation}\label{eq:logall}
  J = -\sum_{i }\log P(i|\hat{\bx}_i) - \sum_{i}\sum_{j \neq i}\log(1-P(i|\bx_j)).
\end{equation}

\subsection{Rationale Analysis}
This section gives a detailed rationale analysis about why minimizing Eq.~(\ref{eq:log}) could achieve the augmentation invariant and instance spread-out feature. Minimizing Eq.~(\ref{eq:log}) can be viewed as maximizing Eq.~(\ref{eq:pos}) and minimizing Eq.~(\ref{eq:neg}).

Considering Eq.~(\ref{eq:pos}), it can be rewritten as
\begin{equation}\label{eq:pos_1}
  P(i|\hat{\bx}_i) = \frac{\exp(\bmf_i^T\hat{\bmf}_i/\tau)}{\exp(\bmf_i^T\hat{\bmf}_i/\tau) + \sum\nolimits_{k \neq i} \exp (\bmf_k^T\hat{\bmf}_i/\tau)},
\end{equation}
Maximizing Eq.~(\ref{eq:pos}) requires maximizing $\exp(\bmf_i^T\hat{\bmf}_i/\tau)$ and minimizing $\exp (\bmf_k^T\hat{\bmf}_i/\tau), k \neq i$. Since all the features are $\ell_2$ normalized, maximizing $\exp(\bmf_i^T\hat{\bmf}_i/\tau)$ requires increasing the inner product (cosine similarity) between $\bmf_i$ and $\hat{\bmf}_i$, resulting in a feature that is invariant to data augmentation. On the other hand, minimizing $\exp (\bmf_k^T\hat{\bmf}_i/\tau)$ ensures $\hat{\bmf}_i$ and other instances $\{\bmf_k\}$ are separated. Considering all the instances within the batch, the instances are forced to be separated from each other, resulting in the spread-out property.

Similarly, Eq.~(\ref{eq:neg}) can be rewritten as,
\begin{equation}\label{eq:neg_1}
  P(i|\bx_j) = \frac{\exp(\bmf_i^T\bmf_j/\tau)}{\exp(\bmf_j^T\bmf_j/\tau) + \sum\nolimits_{k \neq j} \exp (\bmf_k^T\bmf_j/\tau)},
\end{equation}

Note that the inner product $\bmf_j^T\bmf_j$ is 1 and the value of $\tau$ is generally small (say 0.1 in the experiment). Therefore, $\exp(\bmf_j^T\bmf_j/\tau)$ generally determines the value of the whole denominator. Minimizing Eq.~(\ref{eq:neg}) means that $\exp (\bmf_i^T\bmf_j/\tau)$ should be minimized, which aims at separating $\bmf_j$ from $\bmf_i$. Thus, it further enhances the spread-out property.

\subsection{Training with Siamese Network }
We proposed a Siamese network to implement the proposed algorithm as shown in Fig. \ref{fig:framework}. At each iteration, $m$ randomly selected image instances are fed into in the first branch, and the corresponding augmented samples are fed into the second branch. Note that data augmentation is also be used in the first branch to enrich the training samples. For implementation, each sample has one randomly augmented positive sample and $2N-2$ negative samples to compute Eq. (\ref{eq:logall}), where $N$ is the batch size.
The proposed training strategy greatly reduces the computational cost. Meanwhile, this training strategy also takes full advantage of relationships among all instances sampled in a mini-batch \cite{cvpr16lifted}. Theoretically, we could also use a multi-branch network by considering multiple augmented images for each instance in the batch.

\section{Experimental Results}

We have conducted the experiments with two different settings to evaluate the proposed method\footnote{Code is available at \url{https://github.com/mangye16/Unsupervised_Embedding_Learning}}. The first setting is that the training and testing sets share the same categories (\textit{seen testing category}). This protocol is widely adopted for general unsupervised feature learning. The second setting is that the training and testing sets do not share any common categories (\textit{unseen testing category}).
This setting is usually used for supervised embedding learning~\cite{cvpr16lifted}. Following~\cite{cvpr18mom}, we don't use any semantic label in the training set. The latter setting is more challenging than the former setting and it could apparently demonstrate the quality of learned features on unseen categories.

\subsection{Experiments on Seen Testing Categories}

We follow the experimental settings in \cite{cvpr18nce} to conduct the experiments on CIFAR-10~\cite{cifar10} and STL-10~\cite{stl10} datasets, where training and testing set share the same categories.
Specifically, ResNet18 network \cite{cvpr16resnet} is adopted as the backbone and the output embedding feature dimension is set to 128.
The initial learning rate is set to 0.03, and it is decayed by 0.1 and 0.01 at 120 and 160 epoch.
The network is trained for 200 epochs.
The temperature parameter $\tau$ is set to 0.1.
The algorithm is implemented on PyTorch with SGD optimizer with momentum. The weight decay parameter is 5$\times 10^{-4}$ and momentum is 0.9.
The training batch size is set to 128 for all competing methods on both datasets.
Four kinds of data augmentation methods (\emph{RandomResizedCrop}, \emph{RandomGrayscale}, \emph{ColorJitter}, \emph{RandomHorizontalFlip}) in PyTorch with default parameters are adopted.

\begin{table}[t]\small
\centering
 \begin{tabular}{l|p{2cm}<{\centering}}
  \hline
  Methods                          & kNN \\\hline
  RandomCNN                          & 32.1 \\
  DeepCluster (10)  \cite{eccv18cluster}             & 44.4\\
  DeepCluster (1000) \cite{eccv18cluster}              & 67.6\\
  Exemplar \cite{pami16exampler}               & 74.5\\
  NPSoftmax \cite{cvpr18nce}             & 80.8\\
  NCE \cite{cvpr18nce}                         & 80.4\\
  Triplet                        & 57.5\\
  Triplet (Hard)                       & 78.4\\\hline
  Ours                        & \textbf{83.6}\\\hline
 \end{tabular}
  \vspace{-0.3cm}
 \caption{\label{tab:cifar}\small{kNN accuracy (\%) on CIFAR-10 dataset. }}
 \vspace{-0.5cm}
\end{table}

Following \cite{cvpr18nce}, we adopt weighted $k$NN classifier to evaluate the performance.
Given a test sample, we retrieve its top-k ($k =200$) nearest neighbors based on cosine similarity, then apply weighted voting to predict its label \cite{cvpr18nce}.

\vspace{-0.4cm}
\subsubsection{CIFAR-10 Dataset}
\begin{figure}[t]
  \centering
  \includegraphics[height = 5cm, width = 7cm]{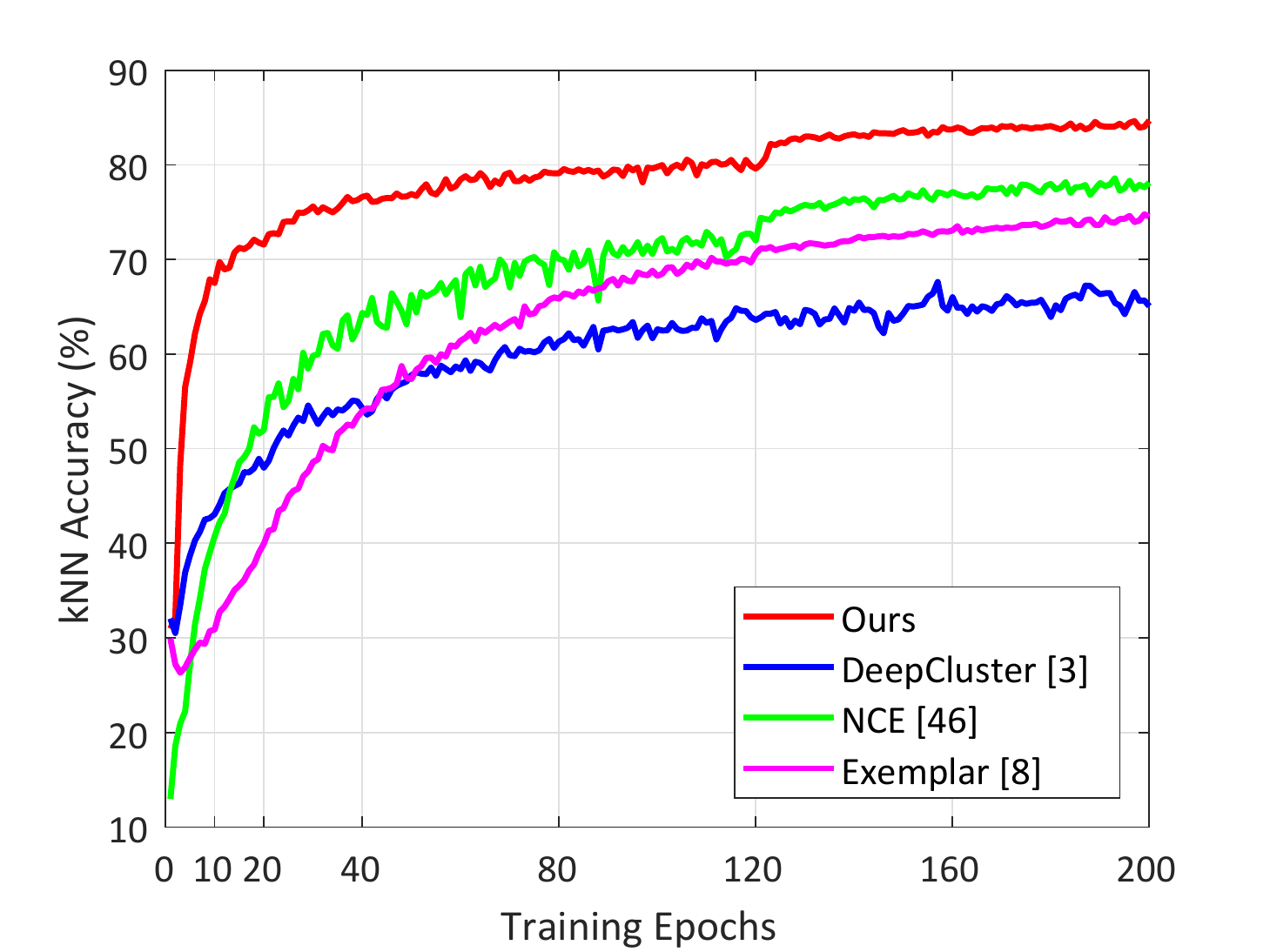}\\
  \vspace{-0.2cm}
   \caption{\small{Evaluation of the training efficiency on CIFAR-10 dataset. kNN accuracy (\%) at each epoch is reported, demonstrating the learning speed of different methods.}}
\label{fig:speed}
\vspace{-0.5cm}
\end{figure}

CIFAR-10 datset \cite{cifar10} contains 50K training images and 10K testing images from the same ten classes. The image size are $32 \times 32$. Five methods are included for comparison: DeepCluster \cite{eccv18cluster} with different cluster numbers, Exemplar CNN \cite{pami16exampler}, NPSoftmax \cite{cvpr18nce}, NCE \cite{cvpr18nce} and Triplet loss with and without hard mining. Triplet (hard) is the online hard negative sample within each batch for training \cite{arxiv17triplet}, and the margin parameter is set to 0.5. DeepCluster \cite{eccv18cluster} and NCE \cite{cvpr18nce} represent the state-of-the-art unsupervised feature learning methods. The results are shown in Table \ref{tab:cifar}.

\textbf{Classification Accuracy.} Table \ref{tab:cifar} demonstrates that our proposed method achieves the best performance (83.6\%) with kNN classifier.
DeepCluster \cite{eccv18cluster} performs well in learning good ``intermediate" features with large-scale unlabelled data, but the performance with kNN classification drops dramatically. Meanwhile, it is also quite sensitive to cluster numbers, which is unsuitable for different tasks.
Compared to Exemplar CNN~\cite{pami16exampler} which uses the classifier weights for training, the proposed method outperforms it by 9.1\%.
Compared to NPSoftmax~\cite{cvpr18nce} and NCE~\cite{cvpr18nce}, which use memorized feature for optimizing, the proposed method outperform by 2.8\% and 3.2\% respectively. The performance improvement is clear due to the idea of directly performing optimization over feature itself.
%
%
%
Compared to triplet loss, the proposed method also outperforms it by a clear margin. The superiority is due to the hard mining nature in Softmax function.

\textbf{Efficiency.} We plot the learning curves of the competing methods at different epochs in Fig. \ref{fig:speed}.
The proposed method takes only 2 epochs to get a kNN accuracy of 60\% while \cite{cvpr18nce} takes 25 epochs and \cite{pami16exampler} takes 45 epochs to reach the same accuracy. It is obvious that our learning speed is much faster than the competitors.
The efficiency is guaranteed by directly optimization on instance features rather than classifier weights \cite{pami16exampler} or memory bank \cite{cvpr18nce}.

\begin{table}[ht]\small
\centering
 \begin{tabular}{l|p{1cm}<{\centering}|p{1.2cm}<{\centering}|p{1.2cm}<{\centering}}
  \hline
  Methods                           & Training          & Linear & kNN \\\hline
  RandomCNN                          & None             & -    &22.4 \\ \hline
  k-MeansNet$^{*}$ \cite{nips11kmeans}     & 105K     & 60.1  & -   \\
  HMP$^{*}$  \cite{er13hmp}                 & 105K     & 64.5  & -   \\
  Satck$^{*}$  \cite{arxiv15stack}          & 105K     & 74.3  & -   \\
  Exemplar$^{*}$  \cite{pami16exampler} & 105K     & 75.4  & -   \\ \hline
  NPSoftmax \cite{cvpr18nce}          & 5K     & 62.3 & 66.8   \\
  NCE  \cite{cvpr18nce}                      & 5K     & 61.9 & 66.2  \\
  DeepCluster(100) \cite{eccv18cluster}     & 5K     & 56.5 & 61.2   \\\hline
  Ours                        & 5K    & 69.5     & 74.1   \\
  Ours                     & 105K    & \textbf{77.9} & \textbf{81.6}   \\ \hline
 \end{tabular}
 \vspace{-0.2cm}
 \caption{\label{tab:stl10}\small{Classification accuracy (\%) with \textit{linear classifier} and \textit{kNN classifier} on STL-10 dataset. $^{*}$Results are taken from \cite{iccv17scat}, the baseline network is different. }}
 \vspace{-0.5cm}
\end{table}
\vspace{-0.2cm}
\subsubsection{STL-10 Dataset}
STL-10 dataset \cite{stl10} is an image recognition dataset with colored images of size $96 \times 96$, which is widely used in unsupervised learning. Specifically, this dataset is originally designed with three splits:
1) \emph{train}, 5K labelled images in ten classes for training,
2) \emph{test}, 8K images from the same ten classes for testing,
3) \emph{unlabelled}, 100K unlabelled images which share similar distribution with labelled data for unsupervised learning.
We follow the same experimental setting as CIFAR-10 dataset and report classification accuracy (\%) with both Linear Classifier (\textit{Linear}) and kNN classier (\textit{kNN}) in Table \ref{tab:stl10}. Linear classifier means training a SVM classifier on the learned features and the labels of training samples. The classifier is used to predict the label of test samples.
We implement NPSoftmax \cite{cvpr18nce}, NCE \cite{cvpr18nce} and DeepCluster \cite{eccv18cluster} (cluster number 100) under the same settings with their released code.
By default, we only use 5K training images without using labels for training.
The performances of some state-of-the-art unsupervised methods (k-MeansNet~\cite{nips11kmeans}, HMP~\cite{er13hmp}, Satck~\cite{arxiv15stack} and Exemplar~\cite{pami16exampler}) are also reported. Those results are taken from \cite{iccv17scat}.

As shown in Table \ref{tab:stl10} , when only using 5K training images for learning, the proposed method achieves the best accuracy with both classifiers (kNN: 74.1\%, Linear: 69.5\%), which are much better than NCE \cite{cvpr18nce} and DeepCluster \cite{eccv18cluster} under the same evaluation protocol.
Note that kNN measures the similarity directly with the learned features and \textit{Linear} requires additional classifier learning with the labelled training data. When 105K images are used for training, the proposed method also achieves the best performance for both kNN classifier and linear classifier.
In particular, the kNN accuracy is 74.1\% for 5K training images, and it increases to 81.6\% for full 105K training images.
The classification accuracy with linear classifier also increases from 69.5\% to 77.9\%.
This experiment verifies that the proposed method can benefit from more training samples.

\vspace{-0.2cm}
\subsection{Experiments on Unseen Testing Categories}

This section evaluates the discriminability of the learned feature embedding when the semantic categories of training samples and testing samples are not overlapped.
We follow the experimental settings described in \cite{cvpr16lifted} to conduct experiments on CUB200-2011(\textit{CUB200}) \cite{cub200}, Stanford Online Product (\textit{Product}) \cite{cvpr16lifted} and Car196 \cite{car196} datasets. No semantic label is used for training.
Caltech-UCSD Birds 200 (\textit{CUB200}) \cite{cub200} is a fine-grained bird dataset.
Following \cite{cvpr16lifted}, the first 100 categories with 5,864 images are used for training, while the other 100 categories with 5,924 images are used for testing.
Stanford Online Product (\textit{Product}) \cite{cvpr16lifted} is a large-scale fine-grained product dataset.
Similarly, 11,318 categories with totally 59,551 images are used for training, while the other 11,316 categories with 60,502 images are used for testing.
Cars (\textit{Car196}) dataset \cite{car196} is a fine-grained car category dataset. The first 98 categories with 8,054 images are used for training, while the other 98 categories with 8,131 images are used for testing.

\textbf{Implementation Details.}
We implement the proposed method on PyTorch. The pre-trained Inception-V1 \cite{cvpr15inception} on ImageNet is used as the backbone network following existing methods \cite{iccv17nca,cvpr16lifted,nips16npair}. A 128-dim fully connected layer with $\ell_2$ normalization is added after the pool5 layer as the feature embedding layer. All the input images are firstly resized to $256 \times 256$. For data augmentation, the images are randomly cropped at size $227 \times 227$ with random horizontal flipping following \cite{cvpr18mom,iccv17nca}.
Since the pre-trained network performs well on CUB200 dataset, we randomly select the augmented instance and its corresponding nearest instance as positive.
In testing phase, a single center-cropped image is adopted for fine-grained recognition as in \cite{iccv17nca}.
We adopt the SGD optimizer with 0.9 momentum. The initial learning rate is set to 0.001 without decay. The temperature parameter $\tau$ is set to 0.1. The training batch size is set to 64.

\begin{table}[t]\small
\centering
 \begin{tabular}{l|cccc|c}
  \hline
  Methods                   & R@1  & R@2   & R@4  & R@8 & NMI      \\\hline
  Initial (FC)              & 39.2 & 52.1  & 66.1 & 78.2  & 51.4\\
  \hline
  & \multicolumn{5}{c}{Supervised Learning} \\
  \hline
  Lifted \cite{cvpr16lifted} & 43.6 & 56.6  & 68.6 & 79.6  & 56.5\\
  Clustering\cite{cvpr17facility} & 48.2 & 61.4 & 71.8 & 81.9 & 59.2\\
  Triplet+ \cite{iccv17smart} & 45.9 & 57.7  & 69.6 & 79.8  & 58.1\\
  Smart+ \cite{iccv17smart}   & 49.8  & 62.3  &74.1 & 83.3 & 59.9 \\
  \hline
    & \multicolumn{5}{c}{Unsupervised Learning} \\
  \hline
  Cyclic \cite{eccv16cycle}   & 40.8 & 52.8  & 65.1 & 76.0  & 52.6\\
  Exemplar \cite{pami16exampler}  & 38.2 & 50.3  & 62.8 & 75.0 & 45.0\\
  NCE \cite{cvpr18nce}            & 39.2 & 51.4  & 63.7 & 75.8  & 45.1\\
  DeepCluster\cite{eccv18cluster} & 42.9 & 54.1  & 65.6 & 76.2  & 53.0\\
  MOM \cite{cvpr18mom}            & 45.3 & 57.8  & 68.6 & 78.4  & 55.0\\\hline
  Ours                            & \textbf{46.2} & \textbf{59.0}  &\textbf{70.1} & \textbf{80.2}  & \textbf{55.4 }\\\hline
 \end{tabular}
 \vspace{-0.3cm}
 \caption{\label{tab:cub}\small{Results (\%) on CUB200 dataset. }}
 \vspace{-0.3cm}
\end{table}

\begin{table}[t]\small
\centering
 \begin{tabular}{l|p{1cm}<{\centering}p{1cm}<{\centering}p{1cm}<{\centering}|p{1cm}<{\centering}}
  \hline
  Methods                & R@1  & R@10   & R@100   & NMI      \\\hline
  Initial (FC)           & 40.8 & 56.7  & 72.1   & 84.0\\\hline
  Exemplar \cite{pami16exampler}  & 45.0 & 60.3  & 75.2   & 85.0\\
  NCE  \cite{cvpr18nce}         & 46.6 & 62.3  & 76.8   & 85.8\\
  DeepCluster\cite{eccv18cluster}  & 34.6 & 52.6  & 66.8   & 82.8\\
  MOM \cite{cvpr18mom}          & 43.3 & 57.2  & 73.2  & 84.4\\\hline
  Ours                          &  \textbf{48.9} & \textbf{64.0}  & \textbf{78.0}  & \textbf{86.0} \\\hline
 \end{tabular}
 \vspace{-0.3cm}
 \caption{\label{tab:stanford}\small{Results (\%) on \textit{Product} dataset. }}
  \vspace{-0.6cm}
\end{table}
\begin{table}[t]\small
\centering
\begin{tabular}{l|cccc|c}
  \hline
  Methods                   & R@1  & R@2   & R@4  & R@8 & NMI      \\\hline
  Initial (FC)              & 35.1 & 47.4  & 60.0 & 72.0  & 38.3\\\hline
  Exemplar \cite{pami16exampler} & 36.5 & 48.1 & 59.2 & 71.0  & 35.4\\
  NCE \cite{cvpr18nce}     & 37.5 & 48.7  & 59.8 & 71.5 & 35.6\\
  DeepCluster\cite{eccv18cluster}  & 32.6 & 43.8  & 57.0 & 69.5  & 38.5\\
  MOM \cite{cvpr18mom}        & 35.5 & 48.2  & 60.6 & 72.4  & \textbf{38.6}\\\hline
  Ours          & \textbf{41.3} & \textbf{52.3}  & \textbf{63.6}   & \textbf{74.9} & 35.8\\\hline
 \end{tabular}
 \vspace{-0.3cm}
 \caption{\label{tab:car}\small{Results (\%) on Car196 dataset. }}
 \vspace{-0.4cm}
\end{table}

\textbf{Evaluation Metrics.} Following existing works on supervised deep embedding learning \cite{iccv17smart,cvpr16lifted}, the retrieval performance and clustering quality of the testing set are evaluated. Cosine similarity is adopted for similarity measurement.
Given a query image from the testing set, $R@K$ measures the probability of any correct matching (with same category label) occurs in the top-$k$ retrieved ranking list \cite{cvpr16lifted}.
The average score is reported for all testings samples. Normalized Mutual Information (NMI) \cite{nmi} is utilized to measure the clustering performance of the testing set.

\textbf{Comparison to State-of-the-arts.}
The results of all the competing methods on three datasets are listed in Table \ref{tab:cub}, \ref{tab:stanford} and \ref{tab:car}, respectively. MOM \cite{cvpr18mom} is the only method that claims for unsupervised metric learning. We implement the other three state-of-the-art unsupervised methods (Exemplar \cite{pami16exampler}, NCE \cite{cvpr18nce} and DeepCluster \cite{eccv18cluster}) on three datasets with their released code under the same setting for fair comparison. Note that these methods are originally evaluated for general unsupervised feature learning, where the training and testing set share the same categories. We also list some results of supervised learning (originate from \cite{cvpr18mom}) on CUB200 dataset as shown in Table \ref{tab:cub}.

Generally, the instance-wise feature learning methods (NCE \cite{cvpr18nce}, Examplar \cite{pami16exampler}, Ours) outperform non-instance-wise feature learning methods (DeepCluster \cite{eccv18cluster}, MOM \cite{cvpr18mom}), especially on \textit{Car196} and \textit{Product} datasets, which indicates instance-wise feature learning methods have good generalization ability on unseen testing categories.
Among all the instance-wise feature learning methods, the proposed method is the clear winner, which also verifies the effectiveness of directly optimizing over feature itself.
Moreover, the proposed unsupervised learning method is even competitive to some supervised learning methods on CUB200 dataset.

%
%
%
\begin{figure*}[ht]
  \centering
  \includegraphics[height = 6.8cm, width = 17cm]{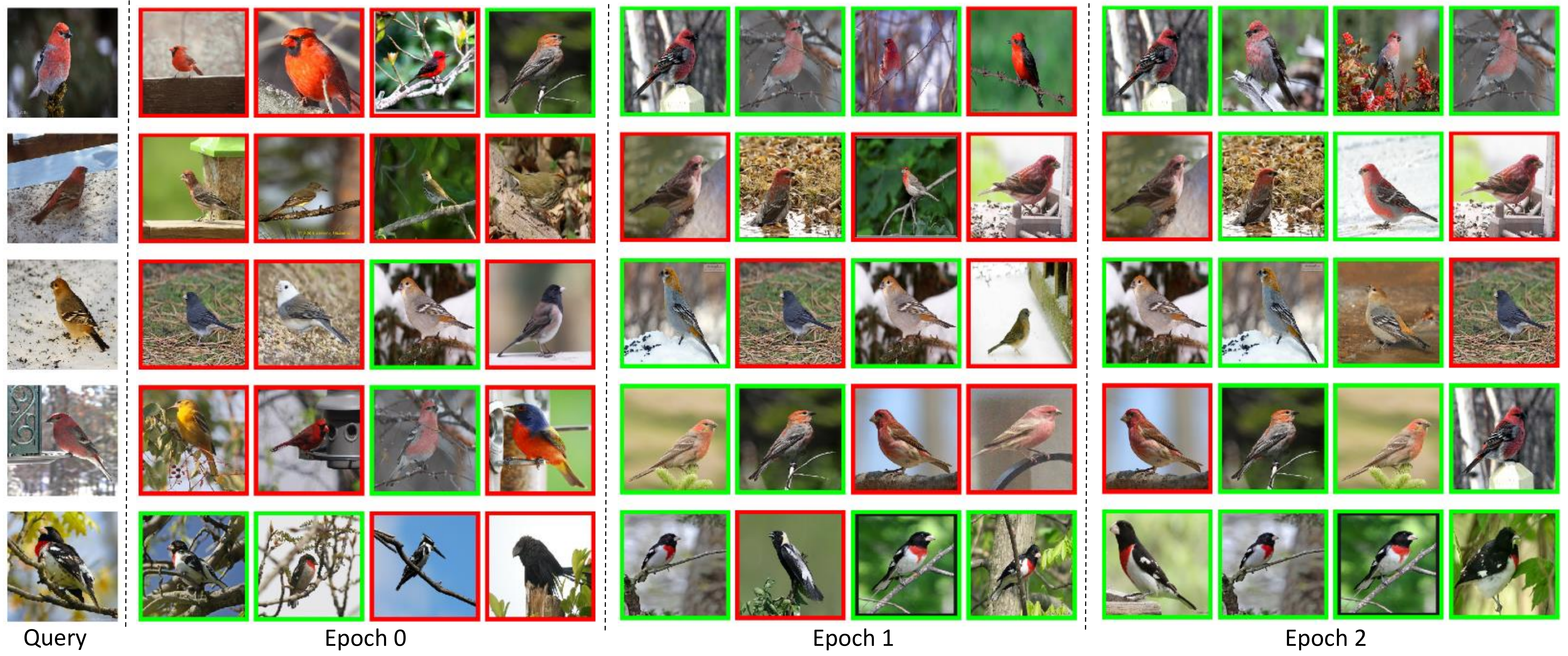}\\
  \vspace{-0.2cm}
  \caption{\small{4NN retrieval results of some example queries on CUB200-2011 dataset. The positive (negative) retrieved results are framed in {\color{green}green} ({\color{red}red}). The similarity is measured with cosine similarity. }}
\label{fig:cub}
\vspace{-0.4cm}
\end{figure*}

\textbf{Qualitative Result.}
Some retrieved examples with cosine similarity on CUB200 dataset at different training epochs are shown in Fig.~\ref{fig:cub}.
The proposed algorithm can iteratively improve the quality of the learned feature and retrieve more correct images.
Although there are some wrongly retrieved samples from other categories, most of the top retrieved samples are visually similar to the query.

\textbf{Training from Scratch.}
%
We also evaluate the performance using a network (ResNet18) without pre-training. The results on the large-scale \textit{Product} dataset are shown in Table \ref{tab:scratch}.
The proposed method is also a clear winner.
Interestingly, MOM \cite{cvpr18mom} fails in this experiment.
The main reason is that the feature from randomly initialized network provides limited information for label mining. Therefore, MOM cannot estimate reliable labels for training.

\begin{table}[t]\small
\centering
\begin{tabular}{l|p{1cm}<{\centering}p{1cm}<{\centering}p{1cm}<{\centering}|p{1.2cm}<{\centering}}
  \hline
  Methods              & R@1  & R@10   & R@100   & NMI      \\\hline
  Random               & 18.4 & 29.4  & 46.0   & 79.8\\\hline
  Exemplar \cite{pami16exampler}  & 31.5 & 46.7  & 64.2   & 82.9\\
  NCE  \cite{cvpr18nce} & 34.4 & 49.0  & 65.2   & 84.1\\
  MOM \cite{cvpr18mom}  & 16.3 & 27.6  & 44.5  & 80.6\\\hline
  Ours                  &  \textbf{39.7} & \textbf{54.9}  & \textbf{71.0}  & \textbf{84.7} \\\hline
 \end{tabular}
 \vspace{-0.2cm}
 \caption{\label{tab:scratch}\small{Results (\%) on \textit{Product} dataset using network without pre-trained parameters. }}
 \vspace{-0.4cm}
\end{table}

\subsection{Ablation Study}
The proposed method imposes two important properties for instance feature learning: data augmentation invariant and instance spread-out. We conduct ablation study to show the effectiveness of each property on CIFAR-10 dataset.

\begin{table}[h]\small
\centering
 \begin{tabular}{l|p{0.8cm}<{\centering}|p{0.8cm}<{\centering}|p{0.8cm}<{\centering}|p{0.8cm}<{\centering}|p{0.8cm}<{\centering}}
  \hline
  Strategy                 & Full   & w/o R & w/o G &  w/o C & w/o F \\\hline
  kNN Acc (\%)              & 83.6  & 56.2  & 79.3   & 75.7 & 82.6\\ \hline
 \end{tabular}
 \vspace{-0.3cm}
 \caption{\label{tab:aug}\small{Effects of each data augmentation operation on CIFAR-10 dataset. 'w/o': Without. 'R': \textit{RandomResizedCrop}, 'G': \textit{RandomGrayscale}, 'C': \textit{ColorJitter}, 'F': \textit{RandomHorizontalFlip}. }}
 \vspace{-0.3cm}
\end{table}
%
\begin{table}[h]\small
\centering
 \begin{tabular}{l|p{1cm}<{\centering}|p{1.1cm}<{\centering}|p{1cm}<{\centering}|p{1cm}<{\centering}}
  \hline
  Strategy                 & Full   & No DA & Hard &  Easy \\\hline
  kNN Acc (\%)              & 83.6  & 37.4  & 83.2   & 57.5\\ \hline
 \end{tabular}
 \vspace{-0.3cm}
 \caption{\label{tab:negative}\small{Different sampling strategies on CIFAR-10 dataset.}}
 \vspace{-0.3cm}
\end{table}

To show the importance of data augmentation invariant property, we firstly evaluate the performance by removing each of the operation respectively from the data augmentation set. The results are shown in Table \ref{tab:aug}. We observe that all listed operations contribute to the remarkable performance gain achieved by the proposed algorithm. In particular, \textit{RandomResizedCrop} contributes the most. We also evaluate the performance without data augmentation (\textit{No DA}) in Table \ref{tab:negative}, and it shows that performance drops significantly from 83.6\% to 37.4\%.
It is because when training without data augmentation, the network does not create any positive concentration property. The features of visually similar images are falsely separated.

To show the importance of spread-out property, we evaluated two different strategies to choose negative samples:
1) selecting the top 50\% instance features that are similar to query instance as negative~(hard negative);
2) selecting the bottom 50\% instance features that are similar to query instance as negative~(easy negative).
The results are shown as ``Hard'' and ``Easy'' in Table ~\ref{tab:negative}.
The performance drops dramatically when only using the \textit{easy negative}. In comparison, the performance almost remains the same as the full model when only using \textit{hard negative}.
It shows that separating hard negative instances helps to improve the discriminability of the learned embedding.

\newcommand{\topcaption}{%
  \setlength{\abovecaptionskip}{-0.5cm}
  \setlength{\belowcaptionskip}{-0.5cm}
  \caption}
\begin{figure}[t]
\begin{minipage}{0.48\linewidth}
\begin{tabular}[b]{p{4cm}<{\centering}}
 \includegraphics[width=4cm,height=2cm]{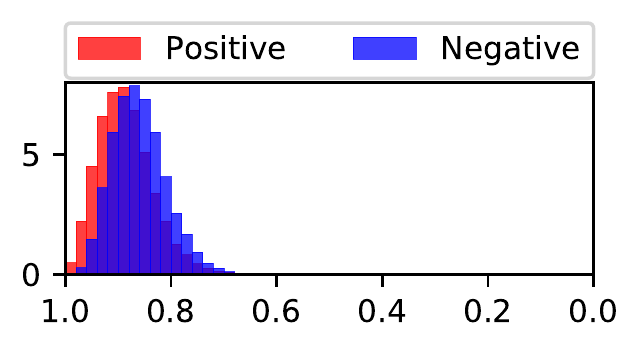}\\
 \vspace{-0.4cm}
 \subcaption*{(a) Random Network}
 \setcounter{subfigure}{0}
 \vspace{-0.2cm}
 \includegraphics[width=4cm,height=2cm]{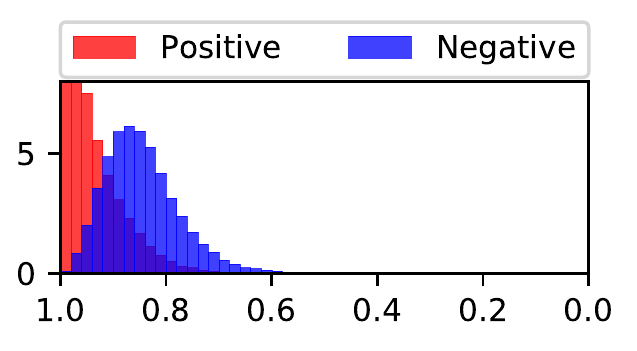}\\ \vspace{-0.5cm}
 \subcaption*{(c) Exemplar \cite{pami16exampler}}
\end{tabular}
\end{minipage}
\begin{minipage}{0.48\linewidth}
\begin{tabular}[b]{p{4cm}<{\centering}}
 \includegraphics[width=4cm,height=2cm]{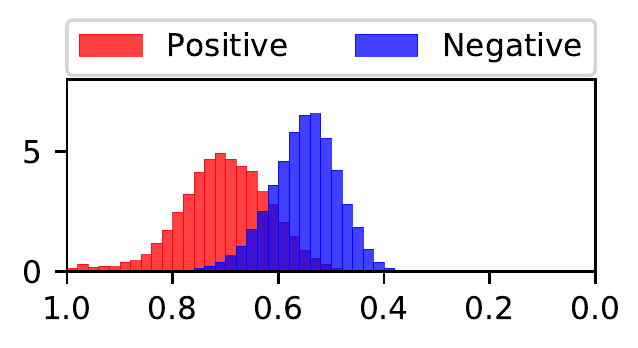}\\
 \vspace{-0.4cm}
 \subcaption*{(b) NCE \cite{cvpr18nce}}
 \vspace{-0.2cm}
 \includegraphics[width=4cm,height=2cm]{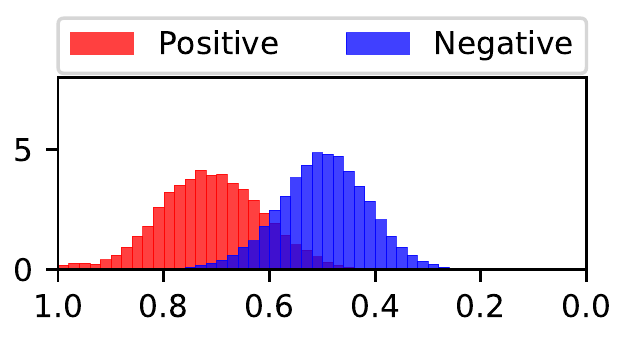}\\ \vspace{-0.5cm}
 \subcaption*{(d) Ours}
\end{tabular}
\end{minipage}
\topcaption{\small{The cosine similarity distributions on CIFAR-10 \cite{cifar10} \label{fig:cifarall} }}
\end{figure}

\subsection{Understanding of the Learned Embedding}
We calculate the cosine similarity between the query feature and its 5NN features from the same category (\textit{Positive}) as well as 5NN features from different categories (\textit{Negative}). The distributions of the cosine similarity of different methods are shown in Fig.~\ref{fig:cifarall}. A more separable distribution indicates a better feature embedding. It shows that the proposed method performs best to separate positive and negative samples. We could also observe that our learned feature preserves the best spread-out property.

\begin{figure}[t]
\begin{tabular}[b]{p{8cm}<{\centering}}
 \includegraphics[width=8cm,height=2cm]{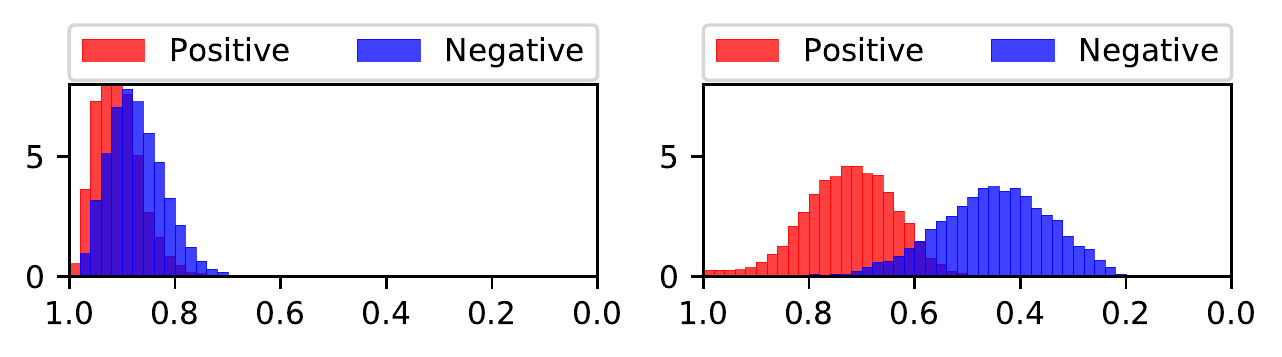}\\ \vspace{-0.5cm}
 \subcaption{Attribute ``\textit{animals vs artifacts}"}
 \vspace{-0.1cm}
  \includegraphics[width=8cm,height=2cm]{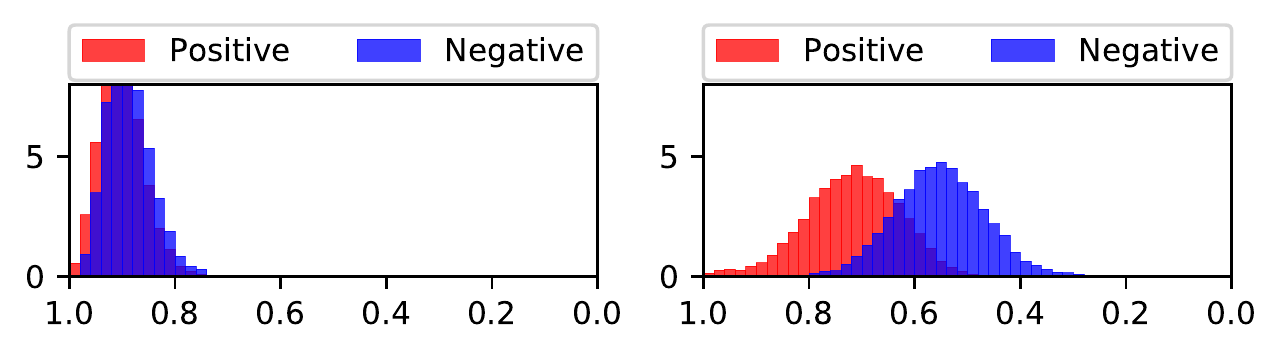}\\
 \vspace{-0.5cm}
 \subcaption{Attribute ``\textit{big vs small shape animal}"}
\end{tabular}
\topcaption{\small{The cosine similarity distributions of randomly initialized network (left column) and our learned model (right column) with different attributes on CIFAR-10 \cite{cifar10}.\label{fig:cifaratt} }}
\end{figure}

It is interesting to show how the learned instance-wise feature helps the category label prediction. We report the cosine similarity distribution based on other category definitions (attributes in \cite{cvpr16unsuper}) instead of semantic label in Fig.~\ref{fig:cifaratt}. The distribution clearly shows that the proposed method also performs well to separate other attributes, which demonstrates the generalization ability of the learned feature.
\section{Conclusion}
In this paper, we propose to address the unsupervised embedding learning problem by learning a data augmentation invariant and instance spread-out feature.
In particular, we propose a novel instance feature based softmax embedding trained with Siamese network, which explicitly pulls the features of the same instance under different data augmentations close and pushes the features of different instances away.
Comprehensive experiments show that directly optimizing over instance feature leads to significant performance and efficiency gains. We empirically show that the spread-out property is particularly important and it helps capture the visual similarity among samples.
%

\vspace{-0.2cm}
\section*{Acknowledgement}
This work is partially supported by Research Grants Council (RGC/HKBU12200518), Hong Kong.
This work is partially supported by the United States Air Force Research Laboratory (AFRL) and the Defense Advanced Research Projects Agency (DARPA) under Contract No. FA8750-16-C-0166. Any opinions, findings and conclusions or recommendations expressed in this material are solely the responsibility of the authors and does not necessarily represent the official views of AFRL, DARPA, or the U.S. Government.
\begin{figure*}[h]
  \centering
  \includegraphics[height = 23cm, width = 17cm]{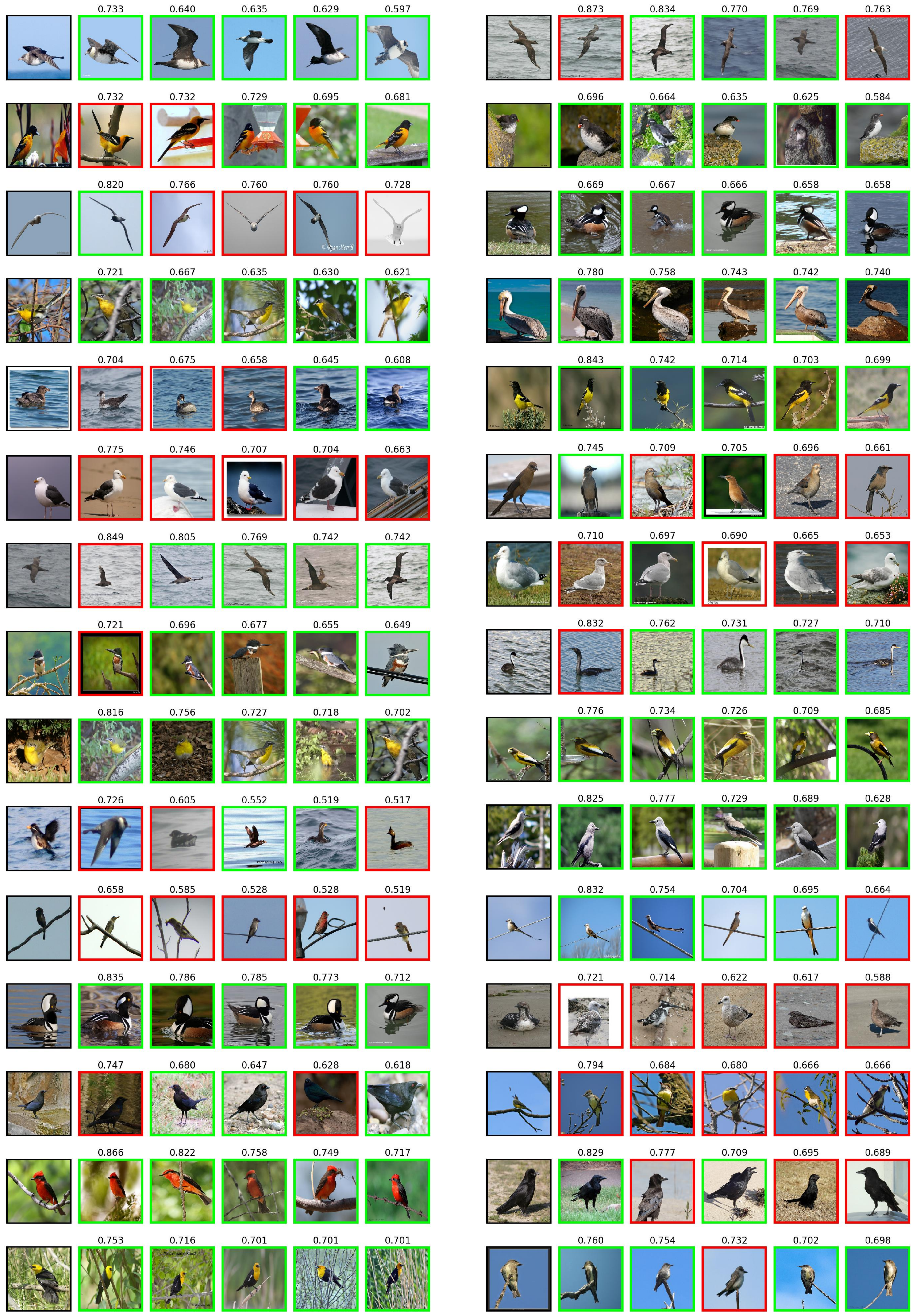}\\
   \caption{\small{Random selected retrieved examples on CUB200 dataset with the learnt instance features. \red{Red} denotes wrong labels and \green{green} represents correct labels. }}
\label{fig:speed}
\end{figure*}

\end{document}